\documentclass[letterpaper]{article} % DO NOT CHANGE THIS
\usepackage{aaai2027}  % DO NOT CHANGE THIS
\usepackage[hyphens]{url}  % DO NOT CHANGE THIS
\usepackage{graphicx} % DO NOT CHANGE THIS
\usepackage{natbib}  % DO NOT CHANGE THIS AND DO NOT ADD ANY OPTIONS TO IT
\usepackage{caption} % DO NOT CHANGE THIS AND DO NOT ADD ANY OPTIONS TO IT
\usepackage{algorithm}
\usepackage{algorithmic}
\usepackage{amsmath}
\usepackage{amssymb}%
\usepackage{newfloat}
\usepackage{listings}
\DeclareCaptionStyle{ruled}{labelfont=normalfont,labelsep=colon,strut=off} % DO NOT CHANGE THIS
\floatstyle{ruled}
\newfloat{listing}{tb}{lst}{}
\floatname{listing}{Listing}

\usepackage{booktabs}

\title{ForeSightGuide: An Anticipatory Framework toward Accurate and Low-Redundancy Guidance for the Visually Impaired}
\author{
Zhiyuan Wang\textsuperscript{\rm 1},
Xu Li\textsuperscript{\rm 1},
Shikang Guo\textsuperscript{\rm 1},
Wei Meng\textsuperscript{\rm 1},
Quan Liu\textsuperscript{\rm 1},
Jie Zuo\textsuperscript{\rm 1}\thanks{Corresponding author. This work was not supported by any organization.}
}

\affiliations{
\textsuperscript{\rm 1}School of Information Engineering, Wuhan University of Technology\\
122 Luoshi Road, Wuhan 430070, China\\
\{wangzy , 357021,  guoshikang\_2022, weimeng,quanliu,zuojie\}@whut.edu.cn
}
\begin{document}

\maketitle

\begin{abstract}
Electronic travel aids are pivotal for the independent mobility of the visually impaired. While Vision-Language Models (VLMs) offer rich environmental understanding, they often suffer from excessive false positives in dynamic scenarios, leading to cognitive overload. To address this, we present ForeSightGuide, an anticipatory assistive guidance framework that couples semantic scene understanding with predictive hazard assessment. Unlike reactive systems, ForeSightGuide leverages the reasoning capabilities of VLMs to anticipate obstacle motion, effectively filtering out non-threatening objects to provide concise, actionable guidance. To validate our approach, we introduce a novel dataset captured in complex, dynamic real-world traffic scenes, designed to benchmark predictive capabilities. Extensive experiments on both public benchmarks and our proposed dataset demonstrate that ForeSightGuide achieves state-of-the-art performance. Notably, it significantly mitigates information overload by reducing redundant alerts to 0.299 per guidance output while maintaining a low missed-hazard rate of 0.112, proving its efficacy for  safe walking assistance.
\end{abstract}


\section{INTRODUCTION}

According to the World Health Organization (WHO), at least 2.2 billion people worldwide have near or distance vision impairment \cite{who:2026}. Visual impairment can reduce walking ability and increase mobility-related safety risks, especially in unfamiliar environments with dense pedestrians and obstacles. Therefore, safe and effective assistive guidance is essential for supporting the independent mobility and social participation of blind and visually impaired persons (BVIPs) \cite{AssistiveGuidance2025}. However, in real-world dynamic scenes, BVIPs still face complex risks caused by moving pedestrians, vehicles, temporary obstacles, and delayed hazard perception.

In recent years, the rapid advancement of sensor technology, artificial intelligence, and computer vision has driven remarkable breakthroughs in electronic travel aids (ETAs), which act as pivotal assistive tools by providing environmental guidance and navigation solutions for BVIPs~\cite{hayamizu2026woofs}. The core technologies of ETA systems for environmental perception, positioning, and navigation can be categorized into non-vision-based technologies and vision-based technology (VBT) \cite{kim:2025uais}. Non-vision-based technologies often rely on pre-deployed dedicated infrastructure, resulting in high costs, poor adaptability to unfamiliar environments, and limited scalability. By contrast, VBT avoids such infrastructure requirements by leveraging wearable or portable visual devices and deep learning algorithms. The evolution from explicit feature extraction to end-to-end models has significantly enhanced ETAs' environmental perception capabilities \cite{patel:2020dar,tapu2020wearable,zhao2026lafgrpo}. Owing to its superior environmental adaptability, strong scalability, and comprehensive perceptual capability, VBT has gradually become a dominant technical route in the research and development of modern ETA systems.

Although existing ETAs have made substantial progress,  safe walking assistance for BVIPs in real-world dynamic scenes remains far from solved. Unlike static indoor navigation or route-level path planning, daily walking often requires users to deal with moving pedestrians, bicycles, vehicles, temporary obstacles, and traffic rules simultaneously. In such situations, an assistive system should not only identify what objects exist in the scene, but also determine whether these objects will affect the user's future path, whether they require immediate attention, and how the warning should be expressed. If a system reports every detected object or only reacts to the current frame, it may either miss future risks or overwhelm the user with unnecessary information. Therefore, practical guidance systems need both semantic scene understanding and predictive hazard assessment.

\begin{figure*}[t]
\centering
\includegraphics[width=0.8\textwidth]{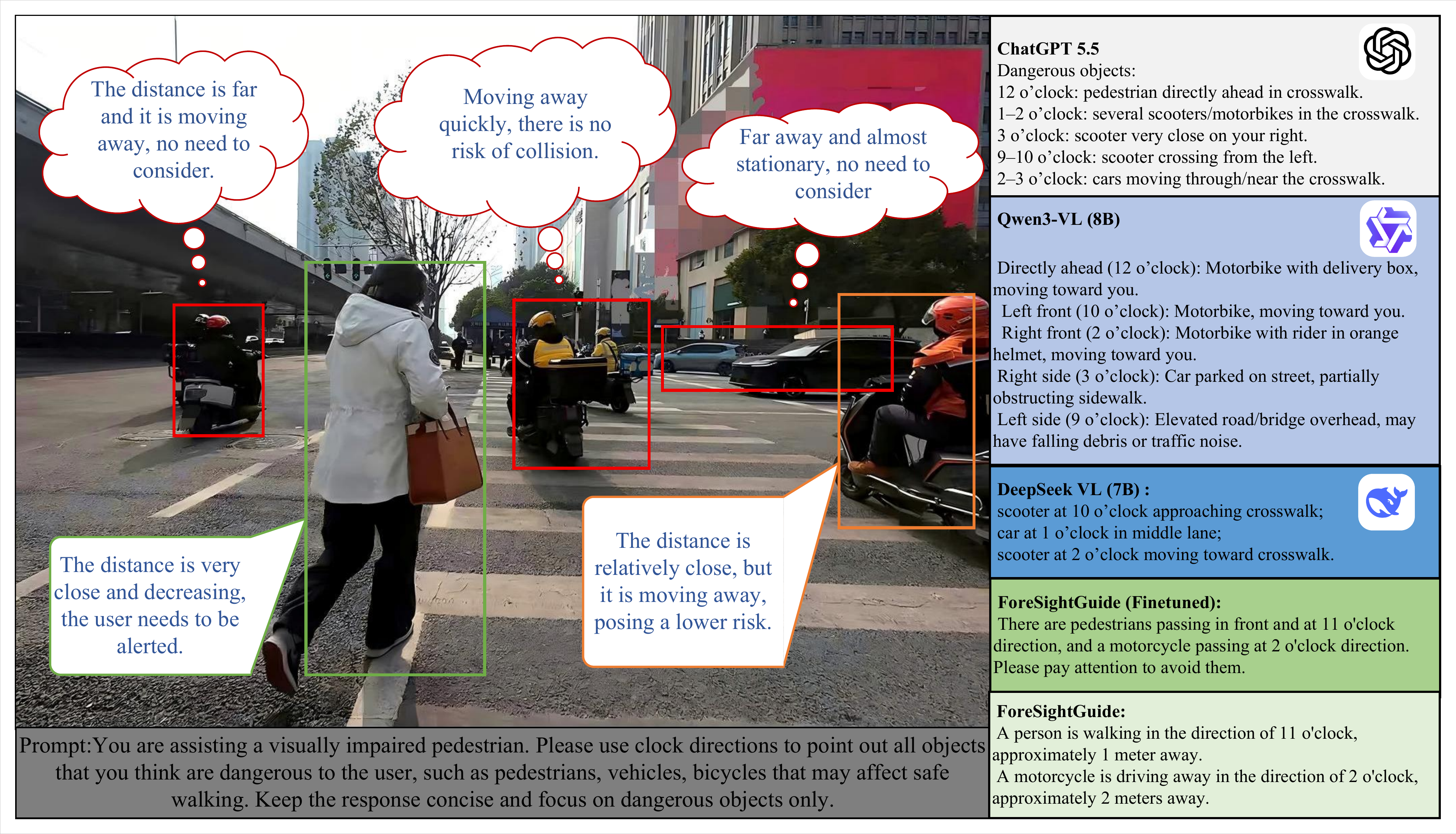} % Reduce the figure size so that it is slightly narrower than the column.
\caption{Example of redundant VLM guidance in a dynamic crossing scene. General-purpose VLMs may report distant or non-threatening objects, which increases cognitive load for BVIPs. ForeSightGuide filters objects by motion and risk, and only reports obstacles that may affect the user's path.}
\label{redundant}
\end{figure*}

The emergence of Vision-Language Models (VLMs) brings new opportunities for assistive guidance for BVIPs. VLMs can convert complex visual information into natural-language descriptions and instructions, helping visually impaired users understand their surrounding environments through accessible verbal feedback. Recent studies have explored VLMs and multimodal language models for assistive navigation, showing their potential in context-aware instruction generation and walking assistance. However, directly applying VLMs to assistive guidance for BVIPs also introduces new risks. Since general-purpose VLMs tend to describe visible content comprehensively,they may produce redundant alerts in dynamic scenes, while recent threat-oriented reasoning work suggests that safety-critical perception should focus on objects that pose real potential risks~\cite{he2026threat}.In Fig. \ref{redundant},we conducted tests on various models under representative outdoor road conditions.For BVIPs, such outputs can increase cognitive load and interfere with timely decision-making. Moreover, VLMs still have limitations in spatial-geometric reasoning and future motion understanding, which may lead to vague or unsafe guidance when obstacles are moving.

To address the core challenges of guidance VLMs for visually impaired users, this paper proposes ForeSightGuide, a novel hybrid fast-slow response guidance framework that integrates precise perception, fast-response safe path planning, and VLM-based deliberative decision-making. In this framework, the safe path planner serves as a rapid reaction layer for time-critical obstacle avoidance, while the VLM functions as a slower reasoning layer for high-level scene understanding and guidance decisions. Specifically, ForeSightGuide extracts semantic information and predicts the future motion trends of obstacles to construct a predictive bird's-eye-view (BEV) map. Based on this map, the system estimates the threat levels of different obstacles and selectively focuses on those that may pose future risks. Unlike purely reactive systems, ForeSightGuide emphasizes motion anticipation and filters out non-threatening or irrelevant objects, thereby reducing redundant alerts while maintaining sensitivity to potential hazards. The reliability of the system is validated with real-world environmental data. Our main contributions are as follows:
\begin{itemize}
\item ForeSightGuide leverages historical visual observations to predict obstacle motion trends instead of relying only on the current frame. It provides the VLM with a dynamic and positional context prefix, making guidance alerts more accurate.
\item Leveraging its predictive capability, ForeSightGuide filters out obstacles that do not pose a real threat through a mask-based mechanism, thereby reducing redundant VLM responses and lowering cognitive load for visually impaired users.
\item ForeSightGuide adopts a hybrid fast-slow response strategy. Our proposed predictive APF method provides high-frequency obstacle avoidance, while the VLM generates low-frequency semantic reminders, combining timely physical guidance with concise verbal feedback.
\end{itemize}

\section{RELATED WORK}

In this section, we review studies related to ForeSightGuide, including electronic travel aids, obstacle-avoidance methods, and VLM-based assistive guidance for visually impaired people.

\paragraph{Electronic Travel Aids.}Traditional ETAs include ultrasonic or infrared sensing devices, smart canes, wearable cameras, haptic feedback devices, and infrastructure-based localization systems \cite{dakopoulos2010wearable,tapu2020wearable}.Representative systems such as the EyeCane provide distance feedback through auditory or haptic signals \cite{maidenbaum2014eyecane}. NavCog uses smartphone localization and Bluetooth beacons to support indoor guidance \cite{ahmetovic2016navcog}. Although these systems are useful for obstacle awareness and localization, they are often limited by semantic understanding, or dependence on pre-installed infrastructure.

\paragraph{Obstacle Avoidance and Path Planning.}
Safe walking assistance for visually impaired people requires reliable route selection and timely obstacle avoidance. Early systems usually relied on graph-search or heuristic planning algorithms to generate safe routes, while recent works further consider semantic maps, user preferences, and dynamic obstacles \cite{Najjar2022Dynamic}. For example, Goswami et al. propose a floor-plan-based localization and guidance system that combines semantic SLAM with A*-based motion command generation \cite{Goswami2024FloorPlanNav}. Surougi and McCann propose MinD, an optimization-based local path-planning method for dynamic outdoor scenes that considers critical moving objects and predicted collision time \cite{Surougi_2023_ICCV}. Na et al. further introduce a Dynamic Walking Corridor Generation method for crowded environments by combining social force models and convex optimization \cite{DynamicCorridor2025}. These methods provide important foundations for safe motion planning.

\paragraph{VLM-Based Assistive Guidance.}
The emergence of VLMs has brought new opportunities for blind walking assistance. By converting complex visual information into natural-language descriptions, VLMs can help visually impaired users perceive their surrounding environments and receive more informative guidance than traditional ETA systems that mainly provide simple alerts or haptic signals. Early work such as SeeWay demonstrated the promise of visual-language fusion for intuitive assistive guidance \cite{Yang2022Seeway}.VLM-based systems are better suited to complex semantic scenes because they can explain what is happening rather than only signaling the presence of an obstacle.

Recent studies have further applied multimodal models to visually impaired assistance in real-world urban scenes.Chao et al. propose an automated context-aware support system for individuals with visual impairment in outdoor urban environments \cite{chao:2025iccvw}. Their method uses VideoLLaMA3 with accessibility-oriented prompting and post-processing to generate spatially and temporally aware guidance. This work shows the potential of VLMs in real-world urban scenes. It also points out that traditional metrics such as BLEU and ROUGE may fail to evaluate navigation instructions that are semantically similar but lexically different.This motivates the use of semantic similarity for evaluating assistive guidance.

WalkVLM is the most relevant prior work to ours because it focuses on VLM-based walking assistance for visually impaired people and introduces the Walking Awareness Dataset (WAD) \cite{yuan:2025walkVLM}. WAD contains 12,000 video-annotation pairs collected from diverse real-world walking scenes.This makes WAD a valuable benchmark for training and evaluating VLM-based walking assistance.WalkVLM adopts CoT-based hierarchical planning and decomposes the task into perception, comprehension, and decision stages. In the perception stage, it uses Detic, an open-vocabulary object detector, to localize scene objects and filters the detected objects by size and confidence score.

Despite these advances, existing VLM-based assistive guidance still has limitations in dynamic scenes. Many methods rely mainly on current observations and do not explicitly predict whether an obstacle will affect the user in the near future. As a result, guidance may become delayed, or the system may describe non-threatening objects and increase cognitive load. Compared with prior path-planning and VLM-based guidance methods, ForeSightGuide emphasizes predictive hazard assessment and concise decision-making. It uses historical visual observations to predict obstacle motion, constructs a predictive BEV map, and filters irrelevant objects before VLM generation. In this way, ForeSightGuide connects risk estimation with semantic guidance,aiming to provide accurate and efficient assistance for BVIPs.

\section{Methodology}

\begin{figure*}[t]
\centering
\includegraphics[width=0.8\textwidth]{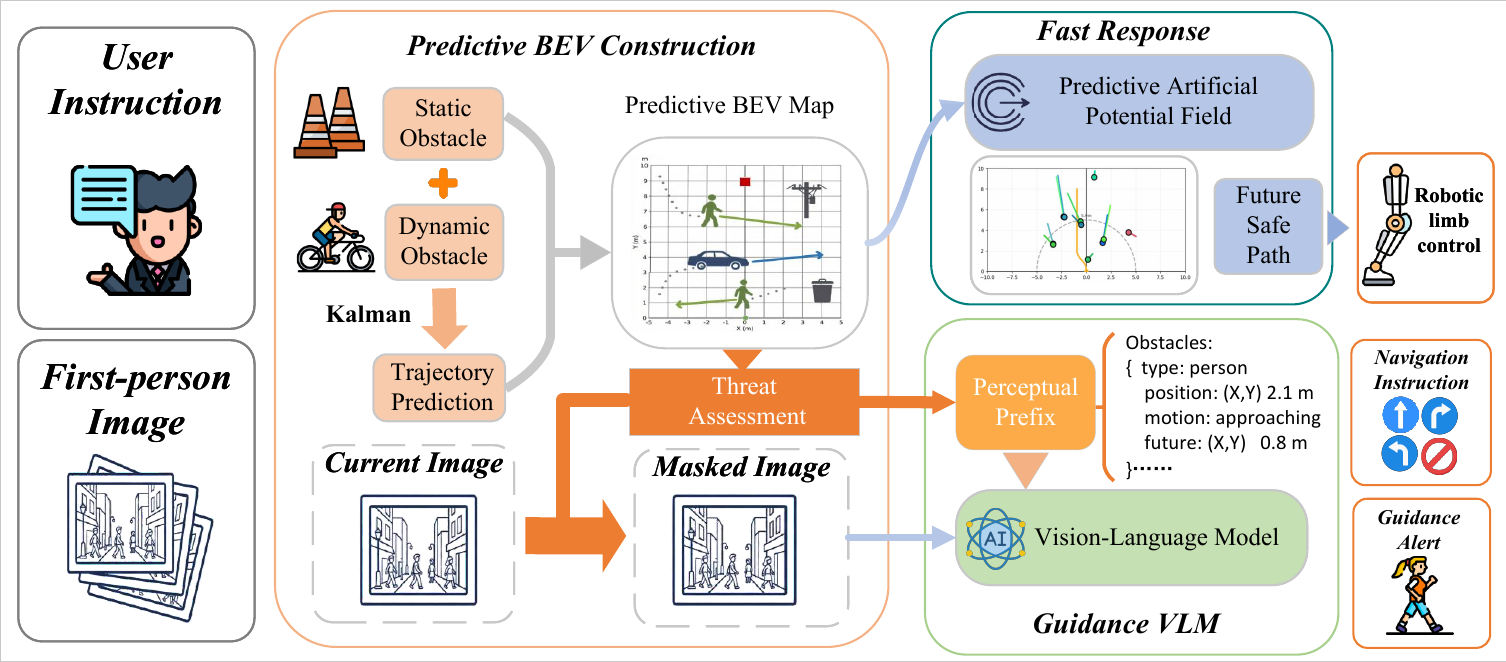} % Reduce the figure size so that it is slightly narrower than the column.
\caption{Overview of ForeSightGuide. The framework predicts obstacle motion, filters non-threatening objects, and provides both high-frequency safe path planning and low-frequency VLM-based guidance for BVIPs.}
\label{overview}
\end{figure*}

ForeSightGuide is a guidance framework designed for BVIPs, as shown in Fig. \ref{overview}. It is built around three stages: environmental perception, safe path planning, and guidance decision-making. Given visual observations, the system combines depth estimation and semantic segmentation to build a predictive bird's-eye-view (BEV) map. Historical trajectories of dynamic obstacles are used to estimate their future motion, allowing the system to reason about potential risks before they occur. Based on this predictive representation, we further propose a predictive artificial potential field method to generate a safe reference path that accounts for the future positions of dynamic obstacles.

To support both real-time control and high-level guidance, ForeSightGuide operates at two decision frequencies. The safe path module runs at a higher frequency and continuously provides reference motion for the ETA. During the lower-frequency VLM decision process, the safe path results are used to control the ETA between VLM outputs. The VLM then generates both guidance alerts and navigation instructions, so that the user receives concise semantic feedback while the ETA still provides timely physical assistance. This design further reduces cognitive load by avoiding frequent verbal updates. In our implementation, the ETA is an electronic extra limb with three degrees of freedom, which provides traction assistance for visually impaired users.
\subsection{Environmental Perception and Risk Filtering}

The environmental perception module provides dynamic scene priors for safe path planning and VLM-based guidance decision-making. For each egocentric image $I_t \in \mathbb{R}^{3\times H\times W}$, we first extract a structured perceptual state:
\begin{equation}
X_t = \mathcal{P}(I_t)
\label{eq:perceptual_state}
\end{equation}
where $\mathcal{P}(\cdot)$ denotes the perception function. Specifically, YOLO11 is used for object detection and instance segmentation, producing object categories, bounding boxes, and instance masks. Depth Pro~\cite{bochkovskiy2025depthpro} is then used to estimate the absolute depth map $Z_t \in \mathbb{R}^{H \times W}$. Based on the camera intrinsic matrix, the RGB-D information is converted into a 3D point cloud and further projected onto a 2D bird's-eye-view (BEV) grid. The resulting perceptual state $X_t$ contains the category, mask, BEV position, and distance of each detected object.

At time $T$, the current egocentric image$I_{\mathrm{curr}}$ is first converted into a structured perceptual state by the perception function $\mathcal{P}(\cdot)$. Conditioned on the historical perceptual states, the module then extracts static and dynamic obstacle information:
\begin{equation}
O_{\mathrm{static}}, T_{\mathrm{dynamic}}
=
\mathcal{Y}\big(I_{\mathrm{curr}} \mid X_{\mathrm{hist}}\big),
\label{eq:obstacle_extraction}
\end{equation}
where $X_{\mathrm{hist}}=\{X_t\}_{t=T-L}^{T-1}$ denotes the structured perceptual states from the previous $L$ frames. The temporal reasoning function $\mathcal{Y}(\cdot)$ performs object association across frames, separates static and dynamic obstacles, and predicts future motion for dynamic obstacles. The outputs $O_{\mathrm{static}}$ and $T_{\mathrm{dynamic}}$ denote the static obstacle set and dynamic obstacle set, respectively. The static obstacle set is defined as
\begin{equation}
O_{\mathrm{static}} =
\{P_s,c_s\},
\label{eq:static_obstacles}
\end{equation}
where $P_s=(x_s,y_s)$ and $c_s$ denote the current BEV position and category of static obstacle $s$.

For each dynamic obstacle $d$, let $\{P_d^t\}_{t=T-L}^{T}$ denote its observed BEV trajectory. A Kalman filter is used to predict future positions and velocities:
\begin{equation}
\{\hat{P}_d^{t'}, \hat{V}_d^{t'}\}_{t'=T+1}^{T+N}
=
f_{\mathrm{kf}}(P_d^{T-L:T}),
\label{eq:motion_prediction}
\end{equation}
where $\hat{P}_d^{t'}$ and $\hat{V}_d^{t'}$ denote the predicted position and velocity at future time $t'$, and $N$ is the prediction horizon. The dynamic obstacle set is then defined as:
\begin{equation}
T_{\mathrm{dynamic}}
=
\left\{
P_d,\,
V_d,\,
\hat{\mathbf{P}}_{d}^{T+1:T+N},\,
\hat{\mathbf{V}}_{d}^{T+1:T+N},\,
c_d
\right\},
\label{eq:dynamic_obstacles}
\end{equation}
where $P_d$, $V_d$, and $c_d$ denote the current position, velocity, and category of dynamic obstacle $d$.

By integrating static obstacles, dynamic obstacles, the module constructs a predictive BEV map:
\begin{equation}
M_{\mathrm{pred}} =
\{O_{\mathrm{static}}, T_{\mathrm{dynamic}}\}
\label{eq:pred_bev}
\end{equation}

Based on the predictive BEV map, we filter the perceived objects before VLM decision-making. The filtering is performed according to the obstacle distance, future position and velocity direction. Static obstacles are retained only if they fall within the predefined hazard range. Dynamic obstacles are retained only if their predicted trajectories enter the user's safety range or move toward the user's walking direction. Objects that remain outside the safety range or move away from the user are treated as non-threatening.

The filtering process is formulated as
\begin{equation}
I_{\mathrm{mask}}, P_{\mathrm{prefix}}
=
\mathcal{F}(M_{\mathrm{pred}}, I_{\mathrm{curr}}),
\label{eq:object_filtering}
\end{equation}
where $\mathcal{F}(\cdot)$ denotes the object filtering function. Given the predictive BEV map $M_{\mathrm{pred}}$, the function first identifies the objects that may affect the user. For objects judged as non-threatening, their instance masks obtained from YOLO11 segmentation are applied to the current image $I_{\mathrm{curr}}$, producing the masked image $I_{\mathrm{mask}}$. For the retained objects, their category, current distance, motion trend, and predicted future position are organized into the perceptual information prefix $P_{\mathrm{prefix}}$.

The retained objects are further ranked by urgency before being sent to the VLM. This process removes irrelevant visual content from the image and provides structured dynamic context, helping the VLM generate more focused guidance.

\subsection{Safe Path Planning} 
Existing obstacle avoidance methods for visually impaired users often rely on the current scene, making them less reliable in dynamic environments.We design a predictive artificial potential field (APF) method that uses the predicted trajectories of dynamic obstacles to generate safe path. The front segment of this path is converted into directional commands, allowing the electronic extra limb to provide physical guidance.

As shown in Algorithm~\ref{alg:papf}, the planner uses a default forward attractive force to provide a basic forward tendency. The attractive gain $\xi$ controls the balance between forward motion and obstacle avoidance: a larger $\xi$ reduces the relative influence of repulsive forces and results in a path with less lateral deviation. Static obstacles and predicted dynamic obstacles generate repulsive forces to push the path away from potential hazards.

\begin{algorithm}[tb]
\caption{Predictive Artificial Potential Field Path Planning}
\label{alg:papf}
\textbf{Input}: Filtered static obstacle set $O^*_{\mathrm{static}}$, filtered dynamic obstacle set $T^*_{\mathrm{dynamic}}$\\
\textbf{Parameter}: Step length $s$, number of candidate repulsion sources $n$, attractive gain $\xi$\\
\textbf{Output}:Planed safe path $Y_{\mathrm{plan}}$
\begin{algorithmic}[1]
\STATE Let $k = 0$.
\STATE Set default forward attractive force $\mathbf{F}_{\mathrm{att}} = \xi \mathbf{e}_{\mathrm{f}}$.
\WHILE{$k < M$}
    \STATE Set $\mathbf{F}_{\mathrm{S}}(\mathbf{q}_k)=0$, $\mathbf{F}_{\mathrm{D}}(\mathbf{q}_k)=0$.
    \FORALL{$s \in O^*_{\mathrm{static}}$}
        \STATE Compute $\mathbf{F}_{\mathrm{s}}(\mathbf{q}_k)$ via Eq.~\ref{eq:Fs}.
        \STATE $\mathbf{F}_{\mathrm{S}}(\mathbf{q}_k) \leftarrow \mathbf{F}_{\mathrm{S}}(\mathbf{q}_k) + \mathbf{F}_{\mathrm{s}}(\mathbf{q}_k)$.
    \ENDFOR
    \FORALL{$d \in T^*_{\mathrm{dynamic}}$}
        \STATE Compute $\mathbf{F}_{\mathrm{d}}(\mathbf{q}_k)$ via Eq.~\ref{eq:force_dynamic}.
        \STATE $\mathbf{F}_{\mathrm{D}}(\mathbf{q}_k) \leftarrow \mathbf{F}_{\mathrm{D}}(\mathbf{q}_k) + \mathbf{F}_{\mathrm{d}}(\mathbf{q}_k)$.
    \ENDFOR
    \STATE Calculate total force: $\mathbf{F}_{\mathrm{total},k} = \mathbf{F}_{\mathrm{att}} + \mathbf{F}_{\mathrm{S}}(\mathbf{q}_k) + \mathbf{F}_{\mathrm{D}}(\mathbf{q}_k)$.
    \STATE Update next path point: $\mathbf{q}_{k+1} = \mathbf{q}_k + s \cdot \frac{\mathbf{F}_{\mathrm{total},k}}{\|\mathbf{F}_{\mathrm{total},k}\|}$.
    \STATE $k \leftarrow k + 1$.
\ENDWHILE
\STATE \textbf{return} $Y_{\mathrm{plan}} = \{q_0, q_1, \dots, q_M\}$.
\end{algorithmic}
\end{algorithm}
\begin{figure}[t]
\centering
\includegraphics[width=0.9\columnwidth]{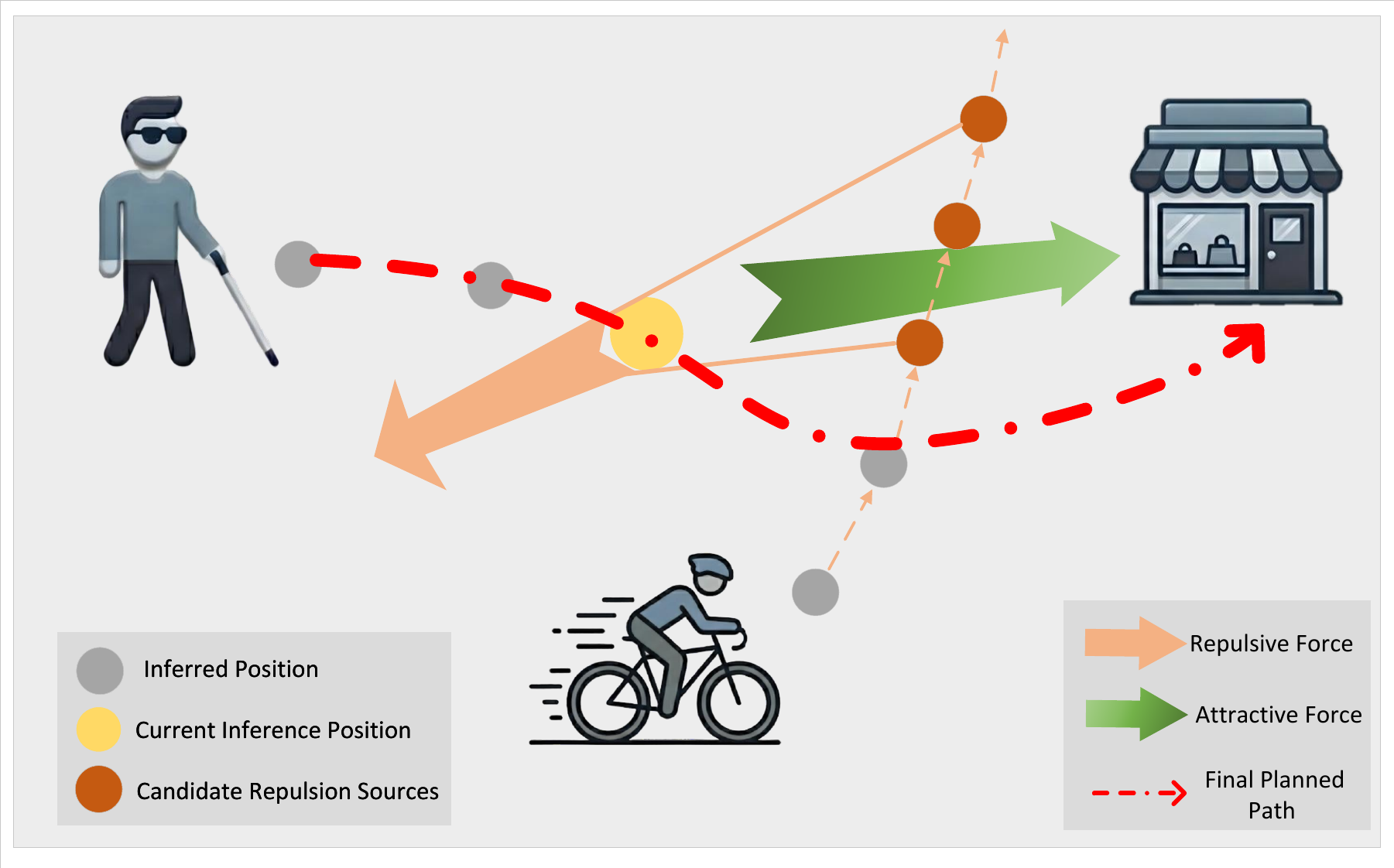} % Reduce the figure size so that it is slightly narrower than the column. Don't use precise values for figure width.This setup will avoid overfull boxes.
\caption{Illustration of the predictive artificial potential field method. The planner uses a default forward attractive force to maintain walking direction, while predicted obstacle positions are treated as candidate repulsion sources to push the reference path away from future hazards.}
\label{apf}
\end{figure}

The planner takes the filtered obstacle sets from the perception module as input, including the static obstacle set $O^*_{\mathrm{static}}$ and the dynamic obstacle set $T^*_{\mathrm{dynamic}}$. 

Static obstacle Repulsive force is generated based on the position of screened static objects, with the repulsive force increasing as the distance decreases:
\begin{equation}
\mathbf{F}_{\mathrm{s}}(\mathrm{q}_k) =  \frac{\mathrm{q}_k - \mathrm{p}_s}{\|\mathrm{q}_k - \mathrm{p}_s\|^3}
\label{eq:Fs} 
\end{equation}
where $\mathrm{p}_s$ is the coordinate on the BEV map of the static object.

The predicted trajectory and speed information are fused to generate repulsive force for approaching dynamic obstacles. As shown in Fig. \ref{apf},  at each iteration, the dynamic repulsion sources are updated according to the predicted obstacle positions at the corresponding future time step.The formula for each obstacle is as follows:

\begin{equation}
\mathbf{F}_d(\mathrm{q}_k) = \frac{1}{n} \cdot  \sum_{\tau=k}^{k+n} \frac{\|\mathbf{V}'_{\tau,d}\|}{\|\mathrm{q}_k - \mathrm{p}'_{\tau,d}\|^3} \cdot (\mathrm{q}_k - \mathrm{p}'_{\tau,d})
\label{eq:force_dynamic} 
\end{equation}
where $n$ represents the number of repulsive sources to consider when calculating repulsion; $\tau$ is the continuous future sequence index starting from current position;$\mathrm{p}'_{\tau,d} $ is the predicted trajectory coordinate of dynamic object $d$ at time $\tau$, aligning repulsion sources with obstacle motion states; $\|\mathbf{V}'_{\tau,d}\|$ is the modulus of dynamic object $d$’s predicted speed at time $\tau$. The summation from $\tau = k$ to $k+n$ constructing a dynamic buffer zone, enhancing capability to avoid approaching obstacle.

The APF step size is temporally aligned with the trajectory prediction interval, defined as $s=v\Delta t$, where $v$ is the typical walking speed of BVIPs and $\Delta t$ is the frame interval. The next path point is then generated as
\begin{equation}
\mathrm{q}_{k+1}
=
\mathrm{q}_k
+
s \cdot
\frac{\mathbf{F}_{\mathrm{total},k}}
{\|\mathbf{F}_{\mathrm{total},k}\|}
\label{eq:dd}
\end{equation}
where $\mathbf{F}_{\mathrm{total},k}$ is the total force at the $k$-th iteration. After reaching the maximum iteration number, the generated reference waypoints are converted into directed line segments for ETA control, providing physical guidance for BVIPs.

\subsection{Guidance VLM}

ForeSightGuide uses Qwen2VL-7B as the VLM backbone to generate high-level guidance for visually impaired users. Direct reasoning on raw images often lacks accurate spatial awareness and future risk perception. As a result, the model may focus on visually salient but non-threatening objects and generate redundant alerts.

To address this issue, ForeSightGuide provides the VLM with the user demand $U$ and two perception-derived inputs: the masked image $I_{\mathrm{mask}}$ and the perceptual information prefix $P_{\mathrm{prefix}}$. The masked image removes non-threatening visual content using instance masks, while the perceptual information prefix provides structured spatial and dynamic cues for the retained obstacles. These inputs enhance the VLM's spatial awareness and motion understanding, leading to more accurate and safety-relevant guidance.

The decision generation process is formulated as
\begin{equation}
O_{\mathrm{vlm}} =
f_{\mathrm{vlm}}(U, I_{\mathrm{mask}}, P_{\mathrm{prefix}}),
\label{eq:vlm_decision}
\end{equation}
where $O_{\mathrm{vlm}}$ denotes the generated guidance output. The output contains both guidance alerts and navigation instructions, which provide concise warnings and actionable walking suggestions for visually impaired users.

\section{Experiments} 

\subsection{Experimental Setup}

\textbf{Dataset}. We evaluate ForeSightGuide on the Walking Awareness Dataset (WAD) and our self-built dynamic-scene dataset. WAD contains 12,000 real-world video-annotation pairs for blind walking assistance, with 1,007 reminder samples for testing~\cite{yuan:2025walkVLM}. Our self-built dataset contains 140 video-annotation pairs collected from complex indoor and outdoor dynamic scenes. Each sample corresponds to one decision-making moment and includes both Guidance Alerts and Navigation Instructions. Initial annotations were generated with GPT-5.4 using the WAD style and then manually refined to ensure accuracy, consistency and suitability for BVIPs.

\textbf{Metrics}. We adopt two types of metrics to evaluate the generated guidance. First, ROUGE-1, ROUGE-2, and ROUGE-L are used as reference text overlap metrics, mainly for comparison with existing VLM-based walking assistance studies. Second, we introduce a semantic similarity metric based on sentence embeddings, because correct guidance may use different wording from the reference annotation. We use \texttt{MiniLM-L12-v2} \cite{wang2020minilm}to encode the generated guidance and ground truth into sentence-level embeddings, and compute their cosine similarity:
\begin{equation}
S_{\mathrm{sem}} =
\cos \left(
\mathbf{e}_{\mathrm{gen}},
\mathbf{e}_{\mathrm{gt}}
\right),
\label{eq:semantic_similarity}
\end{equation}
where $\mathbf{e}_{\mathrm{gen}}$ and $\mathbf{e}_{\mathrm{gt}}$ denote the embeddings of the generated guidance and the ground truth, respectively. Compared with simple word overlap, this metric better reflects whether the generated guidance is semantically consistent with the reference.

\subsection{Experimental Results}

\begin{figure}
    \centering
    \includegraphics[width=0.9\linewidth]{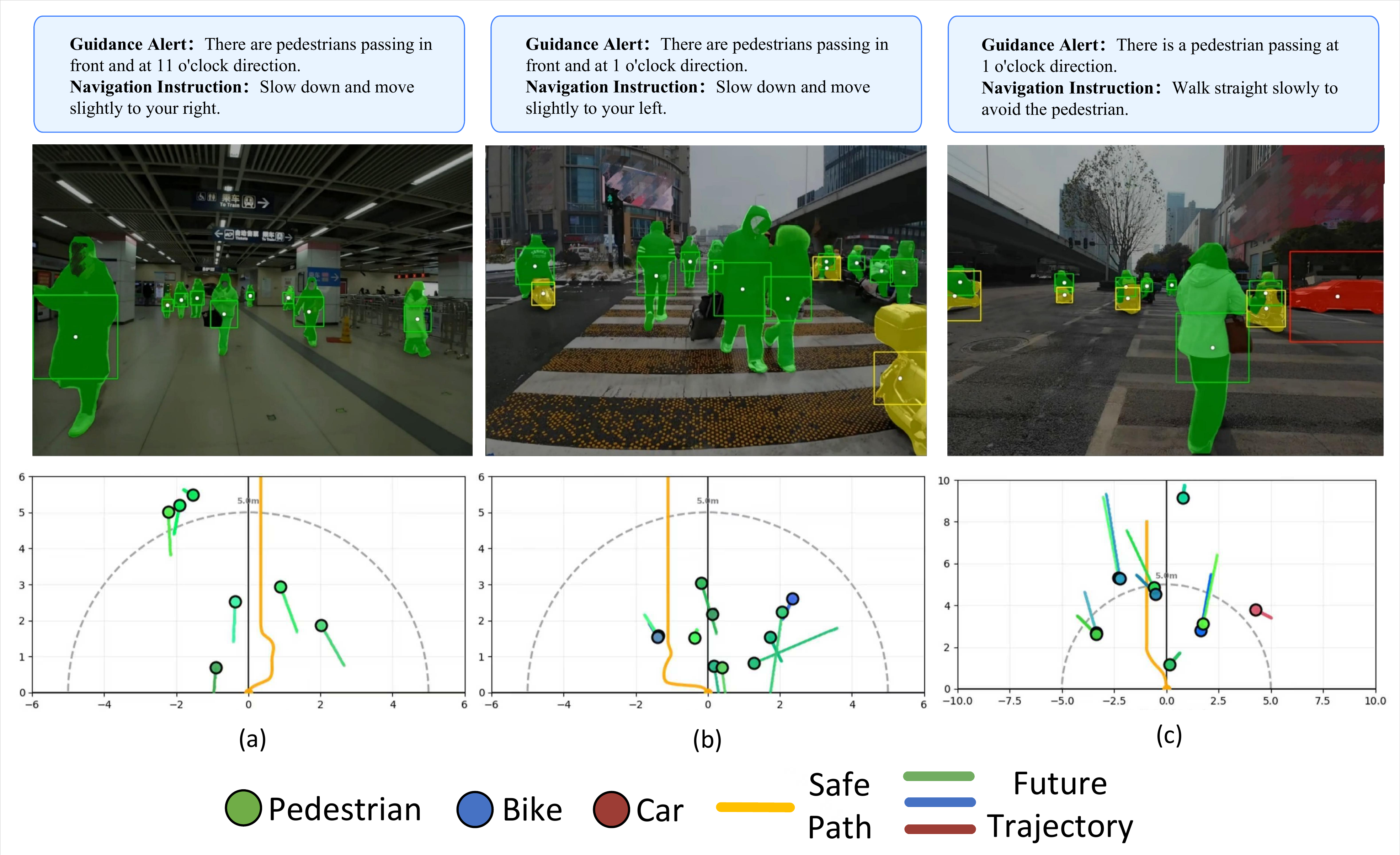}
    \caption{Representative examples of ForeSightGuide in dynamic indoor and outdoor scenes. The upper row shows the generated navigation instructions and guidance alerts, the middle row shows instance-level obstacle perception, and the lower row shows the predictive BEV map with the planned safe path and future obstacle trajectories.}
    \label{fig:examples}
\end{figure}
ForeSightGuide uses Qwen2VL-7B as its backbone.It detects pedestrians, bikes, and vehicles from first-person images, predicts their future trajectories in BEV space, and generates a safe reference path to avoid potential conflicts. It then uses the masked image and perceptual information prefix to guide the VLM in producing concise guidance alerts and navigation instructions. Representative examples in indoor and outdoor dynamic scenes are shown in Fig.~\ref{fig:examples}.

\begin{table*}[t]
\centering
\caption{Quantitative comparison on the WAD dataset and our dataset. The symbol * denotes models fine-tuned on WAD.}
\label{tab:quantitative_comparison}
\small
\setlength{\tabcolsep}{4pt}
\begin{tabular*}{\textwidth}{@{\extracolsep{\fill}}lcccccccc@{}}
\toprule
Model
& \multicolumn{4}{c}{WAD Dataset}
& \multicolumn{4}{c}{Our Dataset} \\
\cmidrule(lr){2-5}
\cmidrule(lr){6-9}
& ROUGE-1 & ROUGE-2 & ROUGE-L & Similarity
& ROUGE-1 & ROUGE-2 & ROUGE-L & Similarity \\
\midrule
Gemma 4 (2B)      & 0.091 & 0.023 & 0.073 & 0.420 & 0.228 & 0.022 & 0.179 & 0.549 \\
DeepSeek (7B)     & 0.100 & 0.010 & 0.074 & 0.468 & 0.220 & 0.023 & 0.164 & 0.525 \\
Qwen3VL (8B)      & 0.101 & 0.013 & 0.072 & 0.488 & 0.233 & 0.032 & 0.183 & 0.571 \\
Qwen2VL (7B)      & 0.117 & 0.023 & 0.089 & 0.485 & 0.239 & 0.033 & 0.182 & 0.558 \\
Qwen2VL (7B)*     & 0.436 & \textbf{0.296} & 0.398 & 0.598 & 0.300 & 0.064 & 0.250 & 0.666 \\
Qwen3VL (8B)*     & 0.415 & 0.261 & 0.372 & 0.593 & 0.380 & 0.133 & 0.278 & 0.731 \\
ForeSightGuide    & \textbf{0.439} & 0.293 & \textbf{0.401} & \textbf{0.611} & \textbf{0.424} & \textbf{0.161} & \textbf{0.329} & \textbf{0.749} \\
\bottomrule
\end{tabular*}
\end{table*}

\paragraph{Comparison with VLM Baselines}. 
We compare ForeSightGuide with general VLMs and WAD-fine-tuned VLMs to evaluate the effectiveness of the proposed perception-guided decision strategy. All fine-tuned models are trained only on the WAD training set, and then evaluated on two test sets: the WAD test set with 1,007 reminder samples and our self-built test set with 140 decision samples. ForeSightGuide uses the open-source Qwen2VL-7B model fine-tuned on WAD as its VLM backbone, without additional fine-tuning on our dataset. The quantitative results are shown in Table~\ref{tab:quantitative_comparison}.

The results show that ForeSightGuide achieves the best overall performance among the WAD-fine-tuned baselines. On WAD, it obtains the highest ROUGE-1, ROUGE-L, and semantic similarity scores, with only a slight decrease in ROUGE-2 compared with Qwen2VL (7B)*. On our self-built dataset, ForeSightGuide outperforms Qwen3VL (8B)*, improving ROUGE-1 from 0.380 to 0.424, ROUGE-2 from 0.133 to 0.161, ROUGE-L from 0.278 to 0.329, and semantic similarity from 0.731 to 0.749. Since ForeSightGuide uses WAD-fine-tuned Qwen2VL-7B and is not further trained on our dataset, the improvement mainly comes from the mask-filtered image and perceptual information prefix, which provide focused visual evidence and spatial-temporal cues for dynamic assistive guidance. It is also worth noting that Qwen3VL (8B)* performs lower than Qwen2VL (7B)* on the WAD test set, but achieves better results on our self-built dataset. This indicates that strong fitting to a single benchmark does not necessarily imply better generalization. Compared with WAD, our dataset further requires the model to generate both guidance alerts and navigation instructions in complex dynamic scenes, where the stronger multimodal reasoning ability of Qwen3VL (8B)* may become more beneficial.
\begin{table}[t]
\centering
\caption{Comparison of redundant alerts and missed important obstacles. The first value denotes the total number of cases, and the value after the slash denotes the average number per decision output.}
\label{tab:redundancy_miss}
\small
\setlength{\tabcolsep}{6pt}
\begin{tabular}{lcc}
\toprule
Model & Redundant & Missed \\
\midrule
GPT-5.5      & 186 / 1.74 & 6 / 0.056 \\
Qwen3VL (8B) & 190 / 1.78 & 42 / 0.392 \\
Ours         & 32 / 0.299 & 12 / 0.112 \\
\bottomrule
\end{tabular}
\end{table}

Table~\ref{tab:redundancy_miss} further evaluates redundant alerts and missed important obstacles. This experiment is conducted on 107 samples selected from our self-built dataset. GPT-5.5 misses the fewest important obstacles, but it produces many redundant alerts, with 186 redundant outputs in total and 1.74 redundant outputs per decision. Qwen3VL (8B) shows a similar redundancy problem and misses more important obstacles. In contrast, ForeSightGuide reduces redundant alerts to 32 in total and 0.299 per decision, corresponding to an 82.8\% reduction compared with GPT-5.5 and an 83.2\% reduction compared with Qwen3VL (8B). Although ForeSightGuide misses slightly more important obstacles than GPT-5.5, it greatly reduces unnecessary verbal feedback while keeping the missed obstacle rate much lower than Qwen3VL (8B). This demonstrates that the proposed masking and predictive perception mechanisms help the VLM focus on truly relevant obstacles, reducing redundant outputs without severely sacrificing safety.
\begin{table}[t]
\centering
\caption{Real-time walking test results.}
\label{tab:real_world_test}
\small
\setlength{\tabcolsep}{6pt}
\renewcommand{\arraystretch}{1.1}
\begin{tabular}{l l}
\toprule
Metric & Result \\
\midrule
Fast response interval & 650 ms \\
Fast response accuracy & 71.32\% (537/753) \\
VLM instruction accuracy & 84.41\% (157/186) \\
Missed important alerts (avg.) & 0.069 \\
\bottomrule
\end{tabular}
\end{table}

\paragraph{Real-time tests.}
We further conduct a real-time walking test to evaluate ForeSightGuide in practical use. In this experiment, users carry a first-person camera and walk through real indoor and outdoor environments without manual intervention. During walking, we record the first-person video frames, fast response results, VLM navigation instructions, and guidance alerts. The fast response module runs continuously, while the VLM generates high-level instructions and reminders every 3 seconds. For evaluation, the recorded scenes and system outputs are provided to ChatGPT, which judges whether each fast response and VLM instruction is correct and whether important guidance alerts are missed. 
As shown in Table~\ref{tab:real_world_test}, the test evaluates 753 fast-response cases and 186 VLM decision cases. The average fast response interval is 650 ms, including image transmission, network delay, and model inference time. The fast response accuracy reaches 71.32\% and the VLM instruction accuracy reaches 84.41\%, while the average number of missed important alerts is 0.069. These results indicate that ForeSightGuide can operate continuously during real walking, but its real-time robustness still has room for improvement.

\begin{table}[t]
\centering
\caption{Ablation study on the perceptual information prefix and mask-based filtering.}
\label{tab:ablation}
\small
\setlength{\tabcolsep}{3pt}
\renewcommand{\arraystretch}{1.08}
\begin{tabular*}{\columnwidth}{@{\extracolsep{\fill}}lcccc}
\toprule
Setting & ROUGE-1 & ROUGE-2 & ROUGE-L & \shortstack[c]{Similarity} \\
\midrule
\shortstack[l]{w/o\\Prefix+Mask} & 0.300 & 0.064 & 0.250 & 0.666 \\
w/o Prefix & 0.353 & 0.129 & 0.288 & 0.673 \\
w/o Mask & 0.383 & 0.154 & 0.320 & 0.682 \\
\textbf{Full Model} & \textbf{0.424} & \textbf{0.161} & \textbf{0.329} & \textbf{0.749} \\
\bottomrule
\end{tabular*}
\end{table}
\paragraph{Ablation Study.}We further conduct ablation experiments on our self-built dataset to evaluate the contributions of the perceptual information prefix and mask-based filtering.As shown in Table~\ref{tab:ablation}, the full model achieves the best performance on all metrics. Removing both the perceptual prefix and the mask-based filtering causes the largest degradation, which indicates that the two components are complementary in improving guidance quality. In particular, removing the perceptual prefix leads to a larger drop than removing the mask alone, suggesting that structured obstacle information contributes more to semantic alignment and instruction generation than visual masking alone. Nevertheless, the mask-based filtering still provides clear gains by suppressing irrelevant visual content and reducing distraction for the VLM. Overall, the ablation results verify the effectiveness of both design choices in ForeSightGuide.
\section{Conclusion}
This paper presents ForeSightGuide, a VLM-based assistive guidance framework for BVIPs in dynamic walking scenes. By using historical visual observations to estimate future obstacle motion, ForeSightGuide provides the VLM with a perceptual information prefix and a mask-filtered image. This helps the VLM focus on safety-relevant objects, improve guidance accuracy, and reduce redundant alerts. Experiments on WAD, our self-built dataset, and real-time walking tests show the effectiveness and preliminary feasibility of the proposed framework. Despite these results, ForeSightGuide still has several limitations. First, the current motion prediction relies on a simple Kalman filter, which is insufficient for modeling complex pedestrian trajectories and social interactions in crowded scenes. Second, the scale of our self-built dataset is still limited, and larger real-world datasets are needed to better evaluate generalization across diverse mobility scenarios. Third, depth estimation introduces considerable computational cost, which affects real-time performance on resource-limited devices. Future work will explore stronger trajectory prediction models, larger-scale data collection, and more efficient perception modules for practical deployment.

\section*{Acknowledgments}
This work was supported in part by the National Natural Science Foundation of China under Grant 52275029 and the Natural Science Foundation of Hubei Province under Grant 2025AFD639 and Grant 2025AFB119, and in part by the Science and Technology Plan Project of Hubei Province under Grant 2025CSA117, and in part by the Wuhan Municipal Natural Science Foundation Exploration Program (Chenguang Program) under Grant 2026040301020038, and in part by the Fundamental Research Funds for the Central Universities.
\bigskip
\noindent 

\bibliography{aaai2027}


\end{document}